# CMRVision: A Foundation Model for Cardiac MR Image Analysis

Athira J. Jacob[1,2], Puneet Sharma[1], and Daniel Rueckert[2,3]

[1] Digital Technology and Innovation, Siemens Healthineers, Princeton, NJ, USA
[2] Chair for AI in Healthcare and Medicine, Technical University of Munich (TUM) and TUM University Hospital, Munich, Germany
[3] Department of Computing, Imperial College London, UK

**Abstract.** Cardiac magnetic resonance (CMR) imaging provides complementary information on cardiac anatomy, function, and tissue characterization across multiple sequences and views. In this work, we investigate foundation model pretraining for 2D CMR and introduce CMRVision, a CMR-specific foundation model trained using DINOv3-style self-supervised learning on a multi-center, multi-sequence cohort of 36 million CMR images. We systematically evaluate architectural and training design choices for domain-specific pretraining. CMRVision is evaluated on two downstream tasks: multi-task segmentation across cine, late gadolinium enhancement (LGE), and mapping sequences, and cine view classification. Our experiments show that CMR-specific pretraining, smaller patch sizes, and patch-level objectives consistently improve downstream performance. Across a multi-task segmentation benchmark, CMRVision achieved the strongest overall performance, outperforming prior natural-image (NI), medical-image, supervised, and CMR foundation model baselines. Improvements were modest but consistent across structures and sequences, with Dice scores ranging from 0.940–0.967 for LV and 0.855–0.905 for myocardium, and reaching 0.929 for RV, 0.920 for LA, and 0.931 for RA. The largest gains were observed for myocardium segmentation in LGE and mapping images. In a zero-shot segmentation task on unseen LGE long-axis views, the model achieved an average Dice score of 0.692, demonstrating cross-view generalization. For cine view classification, CMRVision achieved the highest average accuracy (0.906), compared to prior methods reported in the literature. These results highlight the potential of CMRVision to support robust and generalizable cardiac MRI analysis across multiple sequences and views.

## 1 Introduction

Cardiac magnetic resonance (CMR) is the reference standard for noninvasive assessment of cardiac anatomy, function, and tissue characterization. A typical CMR examination comprises multiple sequences (e.g., cine, late gadolinium enhancement, and parametric mapping) acquired across different anatomical views, each providing complementary clinical information. Consequently, automated analysis requires models that generalize across diverse contrasts, views, and anatomies.

Self-supervised FMs have demonstrated strong transferability in NI vision and are increasingly being explored for medical imaging, including CMR. However, the effectiveness of directly transferring NI representations to CMR remains unclear due to substantial differences in image appearance, acquisition physics, scanner variability, and the subtle anatomical structures encountered in cardiac imaging. While recent studies have explored CMR-specific FMs trained on cine MRI, systematic investigations of FMs spanning multiple CMR sequences remain limited.

We present CMRVision, an FM for cardiac MRI trained on diverse multi-sequence CMR data and adapted from the DINOv3 [23] framework. Motivated by the success of DINO-based SSL in medical imaging [13,18], we systematically investigated architectural and training modifications for CMR representation learning. Our experiments show that CMR-specific pretraining, smaller patch sizes, and patch-level objectives, including Gram-matrix alignment, consistently improve downstream performance. Across a multi-task segmentation benchmark spanning cine, LGE, and mapping sequences, CMRVision achieved the best Dice score in 9 of 11 segmentation targets, and tying for the remaining two, with gains concentrated in LGE, and mapping sequences. In zero-shot segmentation of unseen LGE long-axis views, CMRVision achieved a Dice score 0.692, indicating effective cross-view generalization. For cine view classification, CMRVision achieved an average accuracy of 0.906, outperforming both NI-pretrained models and the previous CMR-specific FM.

## 2 Related Works

**Pretraining:** Several SSL paradigms have been explored: contrastive learning pulls together positive pairs while optionally pushing apart negatives (e.g., DINO [23]); generative methods reconstruct the input to capture its underlying structure (e.g., MAE [9]); and Joint Embedding Predictive Architectures (JEPA) predict embeddings of different views, blending contrastive and generative principles [4]. DINOv3 [23] combines self-distillation losses with patch-level integration and Gram statistics for SoTA results across vision tasks, providing a general-purpose, task-agnostic image encoder.

**Medical Foundation Models:** Many FMs target medical imaging [15], including domain-specific models for chest X-rays [19] and CT [16], and prompt-based FMs such as MedSAM [30]. For CMR, Jacob et al. [13] trained a multi-sequence model via DINO-style self-distillation with benefits across segmentation and classification; CineMA [5] introduced a cine-focused FM with strong cine transfer; and Shad et al. [22] trained a cine video FM with reports for temporal tasks such as EF regression and disease detection. Liu et al. [18] showed NI DINOv3 models transfer well to several medical tasks including cardiac MRI registration [27], but did not evaluate CMR segmentation, leaving dense cardiac modeling understudied. LoRA [12] and related low-rank methods have also been explored as parameter-efficient alternatives to full domain adaptation [10, 17].

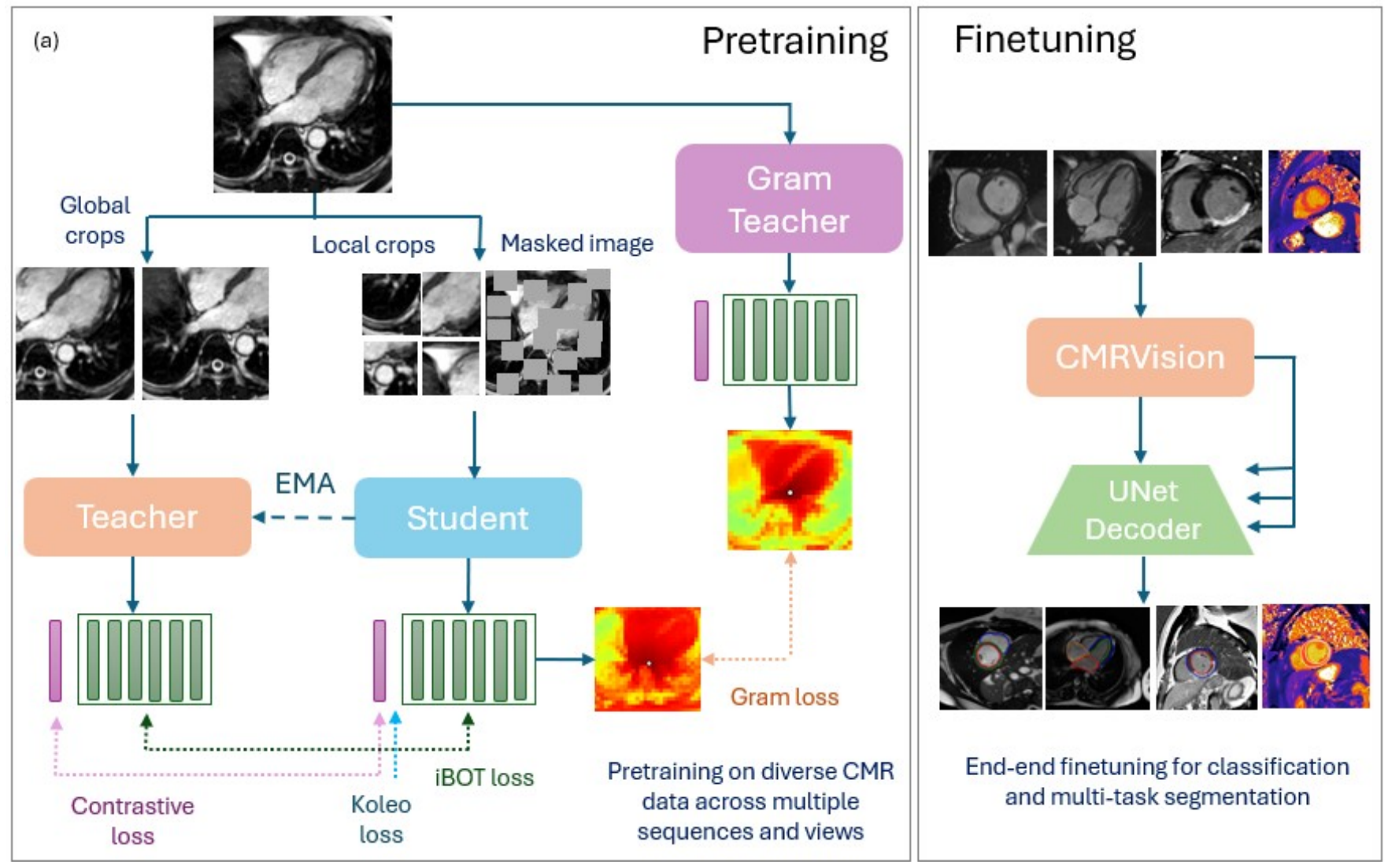

**Fig. 1.** Overview of the pretraining method

**Cardiac Image Analysis:** Domain-specific CMR studies are typically trained and evaluated on small, task-specific datasets. SAX cine segmentation commonly reports Dice of 0.90–0.95 (LV), 0.88–0.91 (myocardium), and 0.90–0.92 (RV) [1, 2]; LAX cine reaches up to 0.94 (LV), 0.88 (myocardium), 0.95 (LA), and 0.96 (RA) on UKBB [1]. For LGE and mapping, 0.84–0.88 (myocardium) and 0.85–0.86 (T1/T2) are reported, often with architecture tailoring and post-processing [14, 28, 31]. Our model matches or exceeds these without task-specific optimization and often with fewer labeled samples.

## 3 Methods

**Pretraining.** We pretrained a ViT-S [26] backbone on the unlabeled CMR dataset (Section 4.1) using a DINOv3-style SSL framework [23] combining four objectives: (i) a DINO loss [3] on global embeddings, enforcing student–teacher consistency across views; (ii) an iBOT patch-level loss [29] aligning local patch embeddings for fine-grained anatomical representation; (iii) KoLeo [6] regularization encouraging a uniform embedding distribution and preventing collapse; and (iv) Gram anchoring [23] aligning second-order patch correlations to preserve CMR-specific inter-patch relationships and spatial coherence. Axial rotary positional embeddings (RoPE) [11] preserved spatial structure across crops. For comparison, a DINOv2-type [20] model was trained with the original three losses (contrastive, iBOT, KoLeo) using mean-centered contrastive loss and learned absolute positional embeddings.

In both cases, we adapt the NI pipeline to CMR via: (i) a smaller patch size (8×8) for fine-grained structures; (ii) early Gram anchoring, enabled by NI initialization; (iii) two-stage training (half- then full-resolution) for efficiency; (iv) domain-specific augmentations, removing operations unsuitable for CMR (e.g., color jitter, vertical flips); and (v) an adjusted crop scale (minimum 0.1) for anatomically meaningful local views. More details are given in Section 4.1.

**Downstream Tasks and Evaluation.** The pretrained encoders were evaluated on two tasks: (a) unified multi-task CMR segmentation in a "segment-all" setting spanning cine short-axis (SAX), cine long-axis (LAX), LGE SAX, and parametric mapping SAX, with no view information provided to the model; and (b) cine view classification. We first compared pretraining strategies under full supervised fine-tuning, then compared the best model against commonly used NI, medical, and supervised baselines. To assess generalization beyond supervised fine-tuning, we also evaluated zero-shot segmentation on LGE LAX images excluded from training; these are harder than cine LAX due to lower SNR, heterogeneous inversion-recovery contrast, and irregular myocardial boundaries from scar enhancement. As the 2-chamber (2CH) view was never seen during training in any sequence, this enables evaluation of both cross-sequence and cross-view generalization. The best model was applied directly to the LGE LAX images without further fine-tuning.

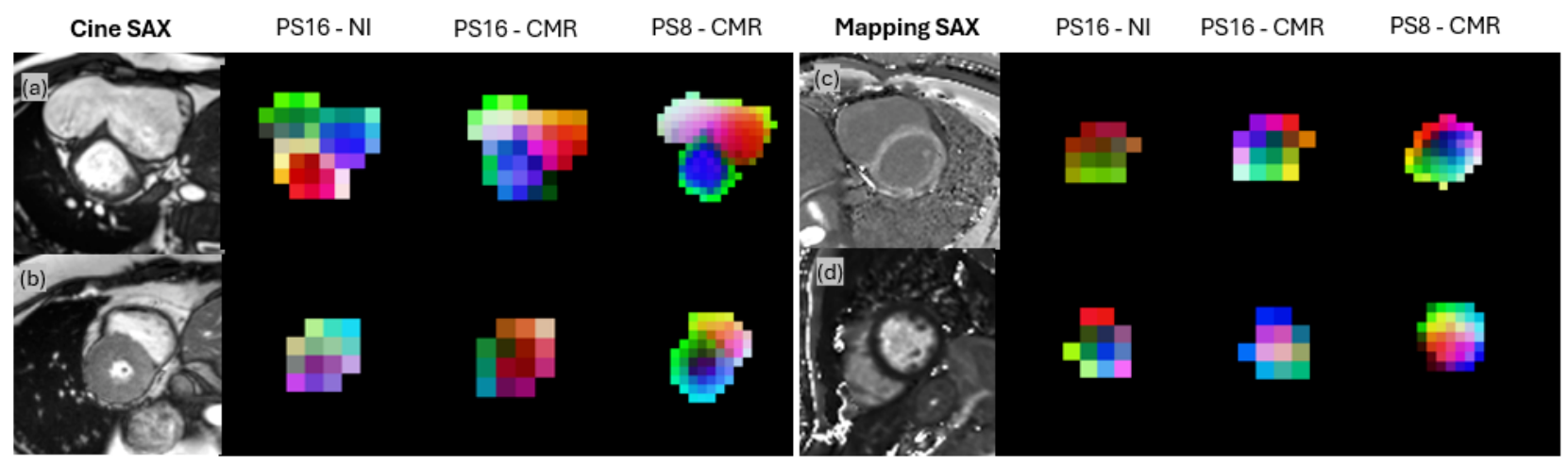

**Fig. 2.** PCA visualizations of the feature space (masked by foreground mask) from pretrained encoders, pretrained on natural images (NI) vs cardiac MRI (CMR) images - a) Cine SAX with dilated RV b) cine SAX with thickened myocardium, c) post-contrast T1 Mapping SAX d) T2 Mapping SAX. PS: Patch size

## 4 Data & Experiments

### 4.1 Pretraining

**Data.** Pretraining used unannotated CMR data from three sources. UK Biobank [24] provided 16,230 participants (1.5 T MAGNETOM Aera; SAX bSSFP cine and three LAX cines). Center 1 contributed 10,073 patients (1.5 T Aera; SAX/LAX cine bSSFP, LGE, T2-weighted, localizer, compressed-sensing, and perfusion). Center 2 provided 1,221 patients (1.5 T and 3 T Avanto/Skyra; SAX/LAX cine bSSFP, LV-outflow and aorta cines, LGE, and first-pass perfusion). In total, the dataset comprised over 36 million images from 27,000 studies on multiple 1.5 T and 3 T scanners, spanning diverse sequences, views, and patient populations for robust FM training.

**Implementation.** Training proceeded in two stages (half- then full-resolution), as larger early-training batch sizes were empirically crucial. Images were resampled to 1 mm in-plane, resized to 224×224, and intensity-normalized (98th-percentile min–max). Multi-crop used 2 global and 6 local crops (112/48 in Stage 1, 224/96 in Stage 2). AdamW followed the schedules in [23], with batch sizes 1024 and 256 in stages 1 and 2. Models were initialized from NI weights; for patch size 8, embeddings were interpolated from 16×16 kernels. Gram teacher updates occurred every 20k iterations. Best models were selected by lowest training loss over 7days, using PyTorch Lightning and Lightly SSL [25].

### 4.2 Finetuning

**Data.** Finetuning experiments used multi-center, annotated CMR data spanning multiple scanners, field strengths (1.5 T and 3T), sequences, and patient populations. Datasets without publicly available annotations were labeled internally by trained annotators. Data for the segmentation task constituted of:

1. Cine SAX (ACDC [16]): LV, myocardium, RV; 150 patients / 300 images (balanced SSFP cine).
2. Cine LAX (Kaggle [18]): LV, myocardium, LA, RA; 436 patients / 874 images, 4-chamber (4CH) view only.
3. LGE SAX (EMIDEC [17] + Centers 1 & 2): LV, myocardium; 1,143 patients / 8,582 images (T1-weighted PSIR and inversion-recovery FLASH).
4. Mapping SAX (Center 3): LV, myocardium; 144 patients / 1,266 images (pre- and post-contrast T1 and T2 mapping).
5. LGE LAX (zero-shot evaluation only, Center 1): myocardium; 22 patients / 57 images, 2-chamber (2CH) and 4CH views.

The downstream segmentation benchmark consisted of the union of datasets 1–4, comprising 1,873 patients and 11,022 images, and zero-shot evaluated on dataset 5. Downstream training and evaluation splits were performed at the patient level. The ACDC, Kaggle, and EMIDEC cohorts were independent of the pretraining data. For the Center 1 and Center 2 cohorts, which also contributed to pretraining, patient-level overlap with the pretraining corpus may exist; however, downstream labels were not used during pretraining. For the ACDC dataset (Cine SAX), we selected one representative basal slice from the end-diastolic and end-systolic frames for each of the 150 patients, resulting in 300 images. This provides a consistent 2D evaluation protocol across datasets with discrete annotated slices while avoiding dataset-specific evaluation procedures.

**Table 1.** Segmentation results on the test set (Dice). **(a)** comparison of pretraining strategies; **(b)** comparison against general/CMR FMs and supervised baselines. Within each block, the best average Dice is in bold and the 2nd best is underlined. All methods are trained in the segment-all setting. † Method required additional class weighting for view-dependent structures for convergence.

| **Method** | **Task Dice** | | | | | | | | | | |
|---|---|---|---|---|---|---|---|---|---|---|---|
| | **Cine SAX** | | | **Cine LAX** | | | | **LGE SAX** | | **Mapping SAX** | |
| | LV | Myo. | RV | LV | Myo. | LA | RA | LV | Myo. | LV | Myo. |
| *a) Comparison of pretraining strategies* | | | | | | | | | | | |
| DINOv3 (S/16, NI) | 0.953 | 0.892 | 0.924 | 0.943 | 0.841 | 0.901 | 0.906 | 0.929 | 0.861 | 0.960 | 0.824 |
| DINOv3/LoRA (S/16, NI) | 0.955 | 0.898 | 0.925 | 0.949 | 0.851 | 0.910 | 0.916 | 0.927 | 0.863 | 0.963 | 0.840 |
| DINOv2 (S/8, CMR) | **0.961** | **0.906** | 0.926 | 0.954 | 0.860 | 0.916 | 0.925 | 0.930 | 0.877 | 0.966 | 0.848 |
| DINOv3 (S/16, CMR) | 0.960 | 0.904 | **0.930** | 0.954 | 0.861 | 0.916 | 0.923 | 0.936 | 0.882 | 0.965 | 0.846 |
| CMRVision (S/8, CMR) | 0.960 | 0.905 | 0.929 | **0.958** | **0.868** | **0.920** | **0.931** | **0.940** | **0.888** | **0.967** | **0.855** |
| *b) Comparison against FMs and supervised baselines* | | | | | | | | | | | |
| SAM2 | 0.819 | 0.742 | 0.857 | 0.811 | 0.361 | 0.890 | 0.883 | 0.840 | 0.723 | 0.804 | 0.455 |
| MedSAM2 | 0.885 | 0.813 | 0.895 | 0.773 | 0.358 | 0.866 | 0.901 | 0.897 | 0.742 | 0.880 | 0.713 |
| Unetr | 0.952 | 0.879 | 0.870 | 0.929 | 0.826 | 0.846 | 0.866 | 0.923 | 0.855 | 0.960 | 0.824 |
| SwinUetr | 0.953 | 0.895 | 0.916 | 0.938 | 0.839 | 0.892 | 0.882 | 0.931 | 0.866 | 0.961 | 0.834 |
| Jacob et al [13] | **0.960** | 0.902 | 0.922 | 0.951 | 0.858 | 0.914 | 0.921 | 0.936 | 0.876 | 0.966 | 0.846 |
| CineMA [5]† | **0.960** | **0.905** | 0.928 | 0.953 | 0.859 | 0.915 | 0.928 | 0.933 | 0.879 | 0.963 | 0.836 |
| CMRVision | **0.960** | **0.905** | **0.929** | **0.958** | **0.868** | **0.920** | **0.931** | **0.940** | **0.888** | **0.967** | **0.855** |

While the LV and myocardium are present across all images and datasets, the atria and RV are inherently view-dependent and therefore appear in substantially fewer samples (e.g., RV in 150 training images and atria in 346, compared to 7,478 for LV and myocardium). This introduces class imbalance in the unified "segment-all" setting, which we retain as it reflects real-world clinical practice, where protocols are oriented toward LV assessment across views and sequences.

The classification benchmark (Center 3) comprised 2,036 images from 147 patients (1,511/218/307 images for train/val/test). View distribution: SAX (41.1%), "other" (23.1%), 2-chamber (11.2%), 4-chamber (9.3%), aorta (8.9%), 3-chamber (6.5%). "Other" covered non-standard views (e.g., 5-chamber) and non-diagnostic images not showing the heart chambers. This dataset was completely independent from the pretraining corpus.

**Table 2.** Cine view classification results on the test set (accuracy). **(a)** comparison of pretraining strategies; **(b)** comparison with other methods. Within each block, the best average accuracy is in bold and the 2nd best is underlined.

| Method | Avg. accuracy | Class | | | | | |
|---|---|---|---|---|---|---|---|
| | | SAX | 2CH | 3CH | 4CH | Aorta | Others |
| *a) Comparison of pretraining strategies* | | | | | | | |
| DINOv3 (ViTS/16, NI) | 0.801 | 0.953 | 0.932 | 0.893 | **1.00** | 0.732 | 0.325 |
| DINOv3/LoRA (ViTS/16, NI) | 0.815 | **0.961** | 0.818 | 0.964 | 0.962 | 0.561 | 0.625 |
| DINOv2 (ViTS/8, CMR) | **0.908** | 0.930 | **1.00** | 0.964 | **1.00** | 0.829 | **0.725** |
| DINOv3 (ViTS/16, CMR) | 0.879 | 0.945 | **1.00** | **1.00** | **1.00** | 0.781 | 0.550 |
| CMRVision (ViTS/8, CMR) | 0.906 | 0.945 | **1.00** | 0.964 | **1.00** | 0.854 | 0.675 |
| *b) Comparison with other methods* | | | | | | | |
| ViTS/8 | 0.707 | 0.813 | **0.705** | 0.929 | **1.00** | 0.195 | 0.600 |
| SwinViT-Tiny | 0.589 | 0.688 | 0.796 | 0.464 | 0.962 | 0.024 | 0.600 |
| Jacob et al [13] | 0.894 | 0.922 | **1.00** | **0.964** | **1.00** | **0.902** | 0.575 |
| CineMA [5] | 0.746 | **0.953** | **1.00** | **0.964** | 0.962 | 0.098 | 0.500 |
| CMRVision (ours) | **0.906** | 0.945 | **1.00** | **0.964** | **1.00** | 0.854 | **0.675** |

**Baselines and Comparisons.** For both tasks, six encoder variants were first evaluated to assess pretraining strategy:

(i) NI DINOv3 ViT-S/16, (ii) LoRA-adapted NI DINOv3 for CMR, (iii) DINOv2 [20] CMR ViT-S/8 (ours), (iv) DINOv3 CMR ViT-S/16 (ours), and (v) DINOv3 CMR ViT-S/8 (CMRVision, ours). The best model was then compared against commonly used general and medical FMs. For segmentation: SAM2 [21] and MedSAM2 [30] (box prompting); UNETR [8] and SwinUNETR [7] (trained from scratch); Jacob et al. [13] (encoder finetuned on the task); and CineMA (publicly available encoder retrained in the segment-all setting on all tasks and structures simultaneously, for fair comparison). For classification: ViTS/8 and SwinViT-Tiny (trained from scratch); Jacob et al. [13] (numbers from the original study); and CineMA [5] (finetuned from the pretrained encoder weights). For the promptable models, the prompts were derived from the ground truth.

**Implementation.** For segmentation, we used a UNETR [8] architecture (ViT-S/8 encoder, U-Net decoder), training a single model in the unified segment-all setting with balanced sampling to account for dataset-size differences. SAM-based models used bounding-box prompting (unlike [13], which used point-prompting), with hollow structures defined as epicardial minus LV masks. For classification, encoders were appended with a 1-layer linear classifier for end-to-end finetuning. Both tasks used full end-to-end finetuning, batch size 64, AdamW (initial learning rate 1e-3), cosine decay over 200 epochs, with best checkpoints by validation Dice (segmentation) or average accuracy (classification).

## 5 Results

**Pretraining Strategy Comparison.** Tables 1a and 2a compare pretraining strategies for multi-task segmentation and cine view classification. For segmentation, CMR-pretrained models consistently outperformed NI-pretrained counterparts, with CMRVision improving average Dice by 1.7 percentage points over NI-pretrained DINOv3 (ViTS/16). Within DINOv3, reducing the patch size from 16 to 8 yielded further gains across most structures and sequences, and domain-specific pretraining outperformed LoRA adaptation, indicating that learning CMR representations directly is more effective than adapting NI features. The largest improvements were in myocardium segmentation and non-cine sequences. PCA visualizations (Fig. 2) further show sharper anatomical boundaries and more coherent representations after CMR-specific pretraining and with the smaller patch size.

For view classification (Table 2a), performance differences between CMR-pretrained models were comparatively small, with all CMR FMs achieving similar accuracy. This suggests that the task relies primarily on global anatomical cues that are already well captured by existing SSL representations, whereas the benefits of CMR-specific pretraining are more pronounced for dense prediction tasks such as segmentation.

**Comparison with Baselines.** Tables 1b and 2b compare CMRVision against supervised, NI, medical-image, and prior CMR FM baselines. CMRVision outperformed the other methods in 9 of 11 tasks and tied for the other two, with particularly large gains over SAM-based models for myocardium segmentation and non-cine sequences. This suggests that diverse multi-sequence pretraining improves robustness for heterogeneous CMR segmentation. For cine view classification (Table 2b), CMRVision achieved an average accuracy of 0.906, the highest among prior methods reported in the literature, and performed comparably to the CMR-pretrained DINOv2 baseline (0.908). Performance was comparable on common SAX and long-axis views, but CMRVision was substantially more accurate on the less frequent "Others" category, driving its superior overall result.

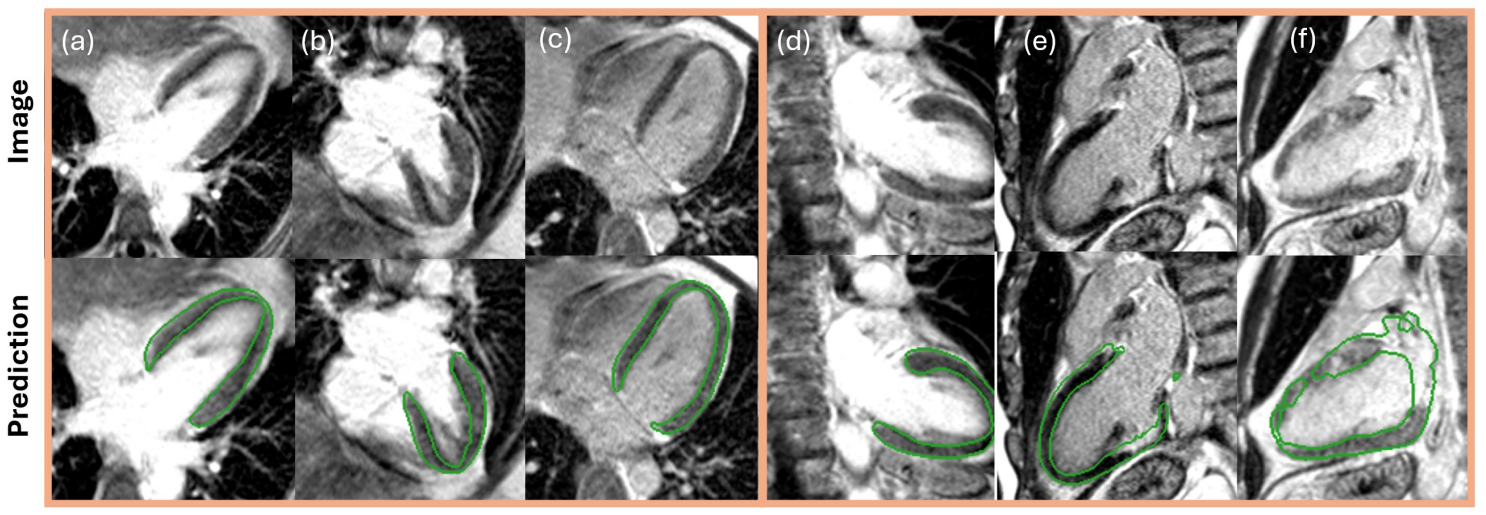

**Fig. 3.** Results from the proposed foundation model on zero-shot segmentation on LGE LAX class. a-c) 4CH, d-f) 2CH.

**Zero-shot evaluation**: To evaluate cross-view generalization, we performed zero-shot segmentation on unseen LGE long-axis myocardium views. CMRVision achieved an average Dice score of 0.692, with scores of 0.612 and 0.794 on 2-chamber and 4-chamber views, respectively, indicating effective transfer to previously unseen anatomical viewpoints. Representative examples (Fig. 3) show that 2CH remain difficult when the imaging plane incompletely captures cardiac chambers, or with anatomical variations.

## 6 Conclusions

This study introduces CMRVision, a cardiac MRI foundation model based on DINOv3-style self-supervised pretraining. CMRVision learns more transferable representations than natural-image-pretrained models, with strong and consistent performance across segmentation, classification, and zero-shot transfer. Our experiments show that patch-level objectives and smaller patch sizes are beneficial, highlighting the importance of domain-specific pretraining design for capturing fine-grained cardiac features. Limitations include the focus on DINO-family objectives, evaluation restricted to CMR, and the absence of an exhaustive comparison of fine-tuning strategies beyond LoRA. Although the downstream benchmark spans 1,873 patients and 11,022 images, some task-specific subsets are relatively small. Patient-level overlap between the pretraining corpus and downstream data from Centers 1 and 2 may also exist, although no downstream labels were used during pretraining. Finally, the current model operates on 2D images and does not explicitly model volumetric or temporal information; extending CMRVision to 3D and spatiotemporal inputs is an important direction for future work. Overall, CMRVision demonstrates the potential of domain-specific self-supervised pretraining to provide a robust, general-purpose backbone for diverse CMR tasks and clinical imaging pipelines across sequences and views.

**Disclaimer**: The concepts and information presented in this paper are based on research results that are not commercially available. Future commercial availability cannot be guaranteed.